\documentclass[lettersize,journal]{IEEEtran}
\usepackage{amsmath,amsfonts}
\usepackage{algorithmic}
\usepackage{algorithm}
\usepackage{array}
\usepackage[caption=false,font=normalsize,labelfont=sf,textfont=sf]{subfig}
\usepackage{textcomp}
\usepackage{stfloats}
\usepackage{url}
\usepackage{verbatim}
\usepackage{graphicx}
\usepackage{cite}
\usepackage{xcolor}
\usepackage{threeparttable}
\usepackage{booktabs}
\usepackage{setspace}
\usepackage{multicol,multirow}

\usepackage[colorlinks,
            linkcolor=red,  
            anchorcolor=blue, 
            citecolor=green,  
            ]{hyperref}

\begin{document}

\title{POEM: Explore Unexplored Reliable Samples to Enhance \\ Test-Time Adaptation}
\author{Chang'an Yi, Xiaohui Deng, Shuaicheng Niu$^*$, and Yan Zhou$^*$
		\IEEEcompsocitemizethanks{
		\IEEEcompsocthanksitem Chang'an Yi, Xiaohui Deng, and Yan Zhou are with the School of Electronic and Information Engineering \& Guangdong Provincial Key Laboratory of Industrial Intelligent Inspection Technology, Foshan University, Foshan, China. Email: \{yi.changan, 2112353032, zhouyan791266\}@fosu.edu.cn 
        \IEEEcompsocthanksitem Shuaicheng Niu is with the College of Computing and Data Science, Nanyang Technological University (NTU), Singapore. E-mail: shuaicheng.niu@ntu.edu.sg
	}}



\maketitle

\begin{abstract}
Test-time adaptation (TTA) aims to transfer knowledge from a source model to unknown test data with potential distribution shifts in an online manner. Many existing TTA methods rely on entropy as a confidence metric to optimize the model. However, these approaches are sensitive to the predefined entropy threshold, influencing which samples are chosen for model adaptation. Consequently, potentially reliable target samples are often overlooked and underutilized. For instance, a sample's entropy might slightly exceed the threshold initially, but fall below it after the model is updated. Such samples can provide stable supervised information and offer a normal range of gradients to guide model adaptation. In this paper, we propose a general approach, \underline{POEM}, to promote TTA via ex\underline{\textbf{p}}loring the previously unexpl\underline{\textbf{o}}red reliabl\underline{\textbf{e}} sa\underline{\textbf{m}}ples. Additionally, we introduce an extra Adapt Branch network to strike a balance between extracting domain-agnostic representations and achieving high performance on target data. Comprehensive experiments across multiple architectures demonstrate that POEM consistently outperforms existing TTA methods in both challenging scenarios and real-world domain shifts, while remaining computationally efficient. The effectiveness of POEM is evaluated through extensive analyses and thorough ablation studies. Moreover, the core idea behind POEM can be employed as an augmentation strategy to boost the performance of existing TTA approaches. The source code is publicly available at \emph{https://github.com/ycarobot/POEM}

\end{abstract}


\section{Introduction}

\begin{figure*}[h]
    \centering  
    \includegraphics[width=\textwidth]{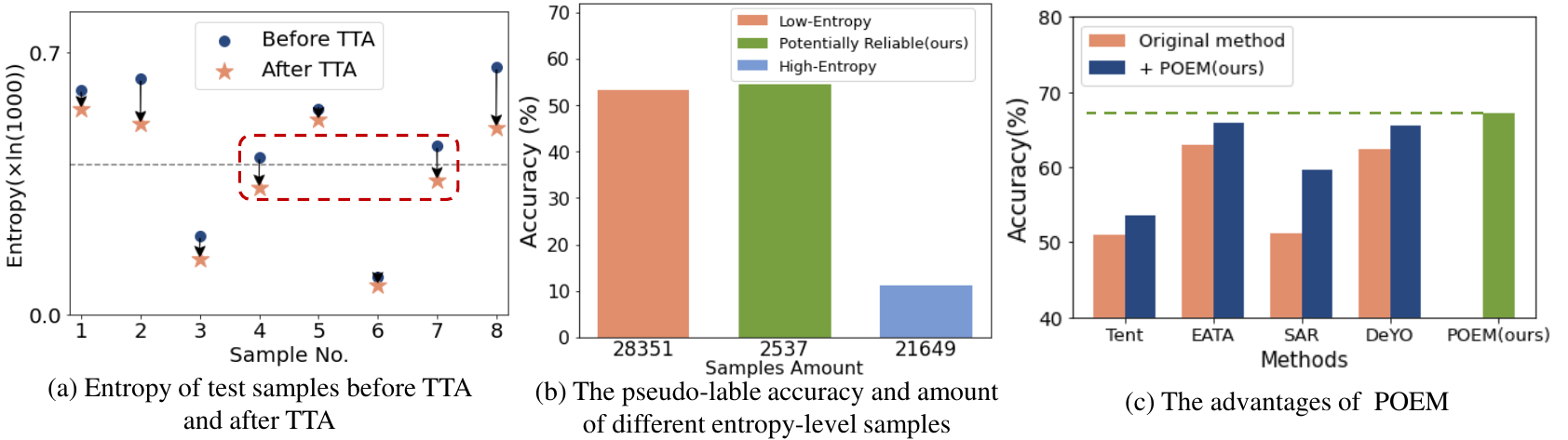}
    \caption{An illustration showing the motivation of this work. (a) Two samples (No. 4, No. 7) are initially predicted as high-entropy but are confirmed as reliable (low-entropy) after TTA. (b) While the number of potentially reliable samples is small, their pseudo-labels exhibit the highest confidence. (c) POEM outperforms existing approaches and its general ideal can also be integrated into these approaches to enhance their performance. The experiments are conducted on the ImageNet-C dataset, which contains 50,000 samples.}
    \label{motifig1}
\end{figure*}

\IEEEPARstart{D}{eep} neural networks (DNNs) perform excellently when the training and test domains follow the same data distribution~\cite{lu2022survey}. However, in real-world applications, domain shifts usually occur since the test data may unavoidably suffer corruptions or natural variations, such as sensor degradation and weather fluctuations~\cite{zhuang2020comprehensive}. In these scenarios, even a well-trained source model may struggle to generalize to the test data. Unlike traditional domain adaptation (DA) tasks, where both labeled source data and unlabeled target data can be accessed simultaneously~\cite{pan2009survey}, target samples in many real-world applications may arrive sequentially. A prominent example is autonomous driving, where continuously evolving environmental conditions lead to ongoing shifts in data distribution that must be addressed in real-time.

Recently, test-time adaptation (TTA) has emerged as a promising paradigm to handle potential domain shifts at test time~\cite{wang2022continual, yuan2023robust, niu2024test, liang2024comprehensive}. Several effective TTA approaches have been proposed, which can generally be categorized into two main types, i.e., Test-Time Training (TTT)~\cite{sun2020test, liu2021TTT++, su2022revisiting} and Fully TTA~\cite{wang2021tent, chen2022contrastive, zhang2023domainadaptor}. While TTT involves modifying the original model training process, this work focuses on Fully TTA (referred to simply as TTA) following Tent~\cite{wang2021tent}. The entropy-based policy is widely employed in existing TTA methodologies~\cite{ niu2022efficient, niu2023towards, lee2024entropy}. These approaches mainly calculate the predicted entropy to select reliable samples, which are then used to adjust the normalization statistics~\cite{ioffe2015batch} to match the test data distribution through entropy minimization. Therefore, the precise selection of reliable samples is crucial for TTA. However, since the entropy of a sample is computed based on the model, that value may change after the model is updated. Existing methods do not re-compute the entropy of a sample if it was initially predicted to have high entropy.

We observe that a sample initially predicted to have high entropy, i.e., higher than the predefined threshold, might later be classified as low entropy (reliable) after TTA (Fig.~\ref{motifig1}(a)). Such samples are not explored in existing entropy-based TTA methods, and we refer to them as potentially reliable or unexplored reliable samples. To investigate them thoroughly, we categorize samples into two groups: low-entropy, and high-entropy. Notably, different samples provide different gradient information in the TTA process. Moreover, some high-entropy samples provide more gradient information compared to other high-entropy samples~\cite{mummadi2021test}. A potentially reliable sample is first a high-entropy sample that, after the model is updated, can drop below the threshold and become low-entropy. Thus, potentially reliable samples are a subset of high-entropy samples. Furthermore, the predicted labels of potentially reliable samples exhibit confidence levels comparable to those of traditional low-entropy samples, yet they significantly surpass the confidence of high-entropy samples (Fig.~\ref{motifig1}(b)).

To further investigate potentially reliable samples for TTA, in this paper, we propose a general approach (\underline{POEM}) to sufficiently ex\underline{\textbf{p}}lore the previously unexpl\underline{\textbf{o}}red reliabl\underline{\textbf{e}} sa\underline{\textbf{m}}ples. In POEM, these newly identified reliable samples can provide stable supervised information and a normal range of gradients to update the model. Additionally, we must strike a balance between preserving the robust representations of existing knowledge and achieving optimal performance on the target data. To this end, we introduce an Adapt Branch network to better capture target information while freezing the higher layers of the source model to maintain transferable knowledge. The advantages of POEM are illustrated in Fig.~\ref{motifig1}(c). In summary, the contributions of this work are as follows:

\begin{itemize}
    \item  We identify and utilize previously unexplored reliable samples, which are overlooked by existing entropy-based approaches, to promote TTA. These samples, initially predicted to have high entropy, can become reliable after the model is updated. Notably, exploring these samples does not significantly increase the computational cost.

    \item We introduce an additional Adapt Branch network designed to strike a balance between robustness and flexibility. Specifically, by integrating its outputs with those of the source model, the overall framework preserves domain-agnostic representations while incorporating fine-grained adjustments tailored to the target data.

    \item  Extensive experiments show that POEM consistently outperforms competing methods across diverse scenarios and environments. Moreover, its two core components can be seamlessly integrated into existing techniques to dramatically boost their performance.

\end{itemize}

\section{Related Works}
\subsection{Test-Time Adaptation}
Test-time adaptation (TTA) constantly fine-tunes a pre-trained source model for unknown test data with distribution shifts~\cite{liang2024comprehensive}. Entropy minimization is commonly employed to adapt the source model in an unsupervised fashion, as entropy is closely related to both error and shift. Tent \cite{Wang_Girshick_Gupta_He_2018} adapts batch normalization (BN) layers by minimizing entropy, where the target model's confidence is assessed based on the entropy of its predictions. EATA~\cite{niu2022efficient} actively selects reliable samples to minimize the loss during inference and introduces a Fisher regularizer to filter out redundant samples, thereby reducing the computational cost. SAR~\cite{niu2023towards} is a reliable and sharpness-aware entropy minimization approach designed to mitigate the effects of noisy test samples with large gradients. Since utilizing extremely low-entropy samples is not enough in TTA, DeYO~\cite{lee2024entropy} addresses the influence of the disentangled factors based on a confidence metric. ROID~\cite{marsden2024universal} aims to tackle the challenging problem of universal TTA. It focuses on promoting consistency in the face of small perturbations to ensure adaptation, while also updating normalization parameters to guarantee efficiency. \cite{liu2024question} employs adaptive entropy minimization based on question types to promote the identification of fine-grained and unreliable instances. \cite{zhang2022memo} can guarantee robustness during test time, by encouraging invariance across augmentations and confident predictions. However, the cross-entropy loss, which is effectively used in classification, is inherently unsuitable for regression problems like pose estimation~\cite{li2021test}. \\
\indent In addition to entropy-based approaches, various other strategies have been proposed to tackle TTA. TT-WA~\cite{liu2025test} fine-tunes the affine coefficients of the BN layers, enabling the model to quickly adapt to different weather scenarios with high reconstruction efficiency. TEA \cite{yuan2024tea} transforms the source model into an energy-based classifier to mitigate domain shifts. AdaContrast ~\cite{chen2022contrastive} combines contrastive learning and pseudo-labeling to address TTA. DomainAdapter~\cite{zhang2023domainadaptor} adaptively merges the training and test statistics within normalization layers. AdaNPC ~\cite{zhang2023adanpc} is a parameter-free TTA approach based on a K-Nearest Neighbor (KNN) classifier, where the voting mechanism is used to attach labels based on $k$ nearest samples from memory. Unlike traditional approaches, CTTA-VDP ~\cite{gan2023decorate} introduces a homeostasis-based prompt adaptation strategy that freezes the source model parameters during the continual TTA process. TTAB ~\cite{zhao2023pitfalls} unveils three pitfalls in prior TTA approaches under classification tasks, based on large-scale open-sourced benchmark approaches and thorough analysis. \\
\indent Since real-time adaptation for pixel-level semantic segmentation is required in tasks such as auto-driving, CoTTA~\cite{wang2022continual} reduces error accumulation based on weight-averaged and augmentation-averaged predictions. DIGA~\cite{wang2023dynamically} is a backward-free segmentation approach that is based on a semantic and a distribution adaptation module, which can adapt the model at both semantic and distribution levels. TTAP finds that classic TTA strategies can not be effectively applied in semantic segmentation tasks~\cite{yi2024question}, it also attempts to disclose the fundamental reasons and propose some possible solutions, such as integrating region-level segmentation and updating the attention module. \\
\indent TTA is also relevant in other fields such as privacy-aware federated learning~\cite{kairouz2021advances}. ATP~\cite{bao2024adaptive} is flexible to handle various kinds of distribution shifts in online federated learning, by adaptively learning the adaptation rates for each target model. FedTHE~\cite{jiang2023test} personalizes federated learning models with robustness to diverse test-time distribution shifts with high computational efficiency. \\

\subsection{Unsupervised Domain Adaptation}
A TTA model adapts knowledge in an online unsupervised manner. However, there are also many other unsupervised domain adaptation solutions. Conventional domain adaptation methods often assume that both source and target data can be accessed simultaneously, nevertheless, this assumption may not be feasible in real-world scenarios due to confidentiality and privacy concerns~\cite{pan2009survey, zhuang2021comprehensive}. A source-free domain adaptation (SFDA) method only utilizes the well-trained source model and unlabeled target data to construct the target domain~\cite{li2024comprehensive}. Knowledge distillation adapts knowledge from a large-scale teacher model to a simple student model~\cite{gou2021knowledge}. However, this kind of cooperation is unidirectional where the teacher model can not be improved. The paradigms of few-shot learning and domain generalization are a bit similar to SFDA. Few-shot DA aims to predict new categories under class inconsistency, instead of addressing the gap of domain distribution~\cite{pal2023domain}. Zero-shot DA aims to capture task-specific knowledge from the source domain when the target data is inaccessible during training~\cite{bian2021domain}. The basic idea of domain generalization is to generalize a model from the source domain to unseen domains~\cite{shu2021open}. However, if the source data is not available due to privacy or storage concerns, a domain generalization approach is not applicable~\cite{wang2022generalizing, zhou2023domain}. \\
\indent Adversarial domain adaptation is based on the idea of generative adversarial networks (GAN)~\cite{goodfellow2020generative}. The basic idea of GAN is to utilize an extra domain discriminator and conduct a min-max game between this discriminator and a feature extractor~\cite{chen2023multicomponent}. Self-supervised learning, based on self-supervised proxy tasks, is a kind of unsupervised learning solution~\cite{gui2024survey}. Contrastive learning has become the most popular self-supervised paradigm, which treats each sample as a category to learn sample-discriminative representations~\cite{zhang2021unleashing}. \cite{wu2024ttagaze} introduces a self-supervised auxiliary task to promote gaze estimation. Compared to cross-entropy-based solutions, a paradigm based on contrastive loss can reduce intra-class feature scattering. In the context of federated learning, federated domain adaptation (FDA) aims to adapt knowledge collaboratively from diverse source domains to related but different unlabeled target domains~\cite{liu2023co}. The difference between an SFDA method and a federated DA method is that the source data is locally kept in the FDA setting. FDAC attempts to address the FDA by thoroughly exploring source models via dual contrastive mechanisms, where both discriminability and transferability are guaranteed~\cite{Yi2025FDAC}. \\
\indent Compared to existing entropy-based TTA approaches, our proposed method, POEM, effectively utilizes the potential of previously overlooked reliable samples, which may transition from high entropy to low entropy after the model is updated. These samples provide stable supervised information and an appropriate range of gradients for model adaptation. The advantages of potentially reliable samples are validated through both empirical analysis and experimental results.

\section{Proposed Method}

\subsection{Problem Statement}

Let $\mathcal{D}^{train} = \{\left(\textbf{x}_i, \textbf{y}_i \right)\}_{i=1}^{N} \in \mathcal{P}^{train}$ be the training data, where $\textbf{x}$, $\textbf{y}$ and $N$ represent the features, labels and data amount, respectively. Let $f_{\theta}$ denote a pre-trained source model with parameters $\theta$. The goal of TTA is to adapt $f_{\theta}$ to the unlabeled test data $\mathcal{D}^{test} = \{\textbf{x}_i\}_{i=1}^{M} \in \mathcal{P}^{test}$ with different data distribution, i.e., $\mathcal{P}^{train}\left(\textbf{x}\right) \neq \mathcal{P}^{test}\left(\textbf{x}\right)$. In a TTA task~\cite{wang2021tent}, a batch of unlabeled test samples arrives at each time step. The model $f_{\theta}$ is updated online, without storing any data. Hence, our goal is to adapt $f_{\theta}$ to fit the target data in an online manner. Despite recent progress, the following challenges still exist: 

1) It is challenging to generate high-quality information to effectively guide the adaptation process. Although pseudo-labels can help filter out noisy samples, accurately selecting
reliable target samples, influencing the quality of the pseudo-labels, requires further exploration. Additionally, the gradients of the selected samples also need more attention.

2) It is challenging to design a framework that consistently extracts core, domain-agnostic information while also maintaining the flexibility necessary to adapt to domain shifts. Although updating normalization layers is a common technique in TTA, whether this technique alone is sufficient to achieve that balance remains an area for further investigation.



\subsection{Motivation}

To tackle the challenges outlined above, two critical questions arise: (1) How can we effectively identify high-quality samples? and (2) Can freezing the higher layers while updating the lower layers strike an optimal balance between extracting domain-agnostic knowledge and adapting to target domain characteristics? As shown in Fig.~\ref{motifig1}, selecting potentially reliable samples offers high-quality supervisory signals—a point that will be discussed in detail later.

Moreover, our experiments indicate that updating only the normalization parameters of the shallow layers achieves performance comparable to updating all layers (Table~\ref{Tent_Shallow_All}). This observation suggests that treating the normalization operations of the entire network uniformly may not be optimal. In fact, as demonstrated by SAR~\cite{niu2023towards} and DeYO~\cite{lee2024entropy}, not updating the normalization parameters in the higher layers does not result in significant GPU time savings (Table~\ref{eff}). Hence, it is crucial to make effective use of these parameters to boost performance. Motivated by these insights, we propose the Adapt Branch network, which collaborates with the source model to strike a balance between extracting general representations and incorporating fine-grained, target-specific adaptations.

\subsection{Overall Framework}

\begin{figure*}[t]
\centering
\includegraphics[width=\textwidth]{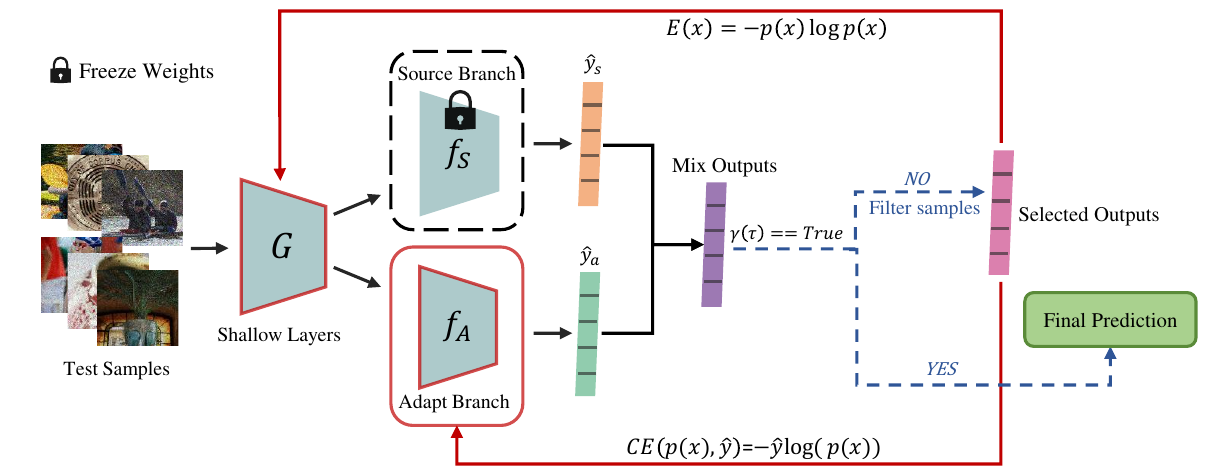} 
\caption{The overall framework of the proposed POEM approach. The selected reliable samples are utilized to update the normalization parameters of both the Shallow Layers and the Adapt Branch network. The selection and update process iterates over some rounds, a hyper-parameter set beforehand.}
\label{overview}
\end{figure*}

Based on the analysis above, we propose a general approach called POEM, as illustrated in Fig.~\ref{overview}. POEM comprises two primary components: \begin{enumerate} \item \textbf{Discovery of Potentially Reliable Samples:} This component focuses on identifying target samples that provide stable and informative gradient signals for updating the model. \item \textbf{Adapt Branch Network:} This network is designed to work in tandem with the source model, seamlessly integrating target-specific knowledge while preserving transferable information. \end{enumerate}

The source model is structured into two parts: the Shallow Layers and the Source Branch network (comprising the higher layers). Notably, the Adapt Branch network mirrors the deep architecture, initialization, and input of the Source Branch network. For each batch of target samples, only the normalization layers within the shallow layers and the Adapt Branch network are updated using selected reliable samples, while the Source Branch network remains frozen. This selective update strategy ensures that the model effectively acquires new target domain insights while retaining essential domain-agnostic information.

\subsection{Identifying Potentially Reliable Samples}
\label{PRSdiscover}
In an entropy-based TTA approach, the objective is to minimize the loss $E$ for each batch of samples to bridge domain gaps through normalization~\cite{wang2021tent}. For each test sample $x$ $\in$ $\mathcal{D}^{test}$, its entropy is computed using:

\begin{equation}\label{eq1}
\begin{aligned}
& E(x) = - \sum_{c=1}^C p^c(x)\log p^c(x) ,\\
\end{aligned}
\end{equation}
where $p(x)$ denotes the model's predicted probability for sample $x$, and $C$ represents the number of categories. After entropy 
computation, the sample $x$ is deemed reliable if $E(x)$ $\leq$ $E_0$, where $E_{0}$ is the predefined entropy threshold~\cite{niu2022efficient}. These reliable samples, characterized as low-entropy, are then utilized to update the model.



However, this entropy-based policy has its drawbacks. Firstly, the results are highly sensitive to the chosen threshold.  Varying the threshold leads to different samples being selected for model parameter updates, which can result in inconsistent outcomes. Second, since the threshold is manually defined, it does not adapt to changes in the model’s parameters. As a result, accurately selecting reliable samples becomes challenging, as the predicted entropy of a sample may change after the model is updated. Therefore, an intelligent mechanism that is robust to manually predefined thresholds is essential.


We observe that some samples, initially predicted as high-entropy, may transition to low-entropy after the model has been updated. Consequently, we categorize the samples into two groups: low-entropy and high-entropy. A potentially reliable sample is defined as follows: it is classified as high-entropy before the model adaptation and low-entropy after the adaptation. As shown in Fig.~\ref{motifig1}(a), following the model update, some samples originally classified as high-entropy tend to shift to low-entropy, while low-entropy samples typically maintain their classification. While distinguishing between low-entropy and high-entropy samples is relatively straightforward, identifying high-entropy samples that may transition to low-entropy (i.e., potentially reliable samples) is more challenging. Potentially reliable samples, which form a subset of high-entropy samples, provide valuable gradient information to supervise the model update \cite{mummadi2021test}. Therefore, enhancing the model's ability to accurately identify these samples will simultaneously improve its overall performance.

Unlike existing entropy-based approaches~\cite{wang2021tent, niu2022efficient, niu2023towards, lee2024entropy, marsden2024universal}, POEM focuses on identifying and utilizing potentially reliable samples within each batch of target data. We propose a novel solution that involves iteratively selecting potentially reliable samples and sequentially updating the model. The number of iterations is a hyper-parameter. In the next section, we will discuss the impact of this hyper-parameter and analyze whether exploring the potentially reliable samples brings much additional computational cost. Let $\theta$ denote the model parameters, we calculate the entropy of each test sample and select the reliable ones using:
\begin{equation}\label{eq2}
\begin{aligned}
    & S_{(t;\theta_{t})} = \cup_{\mathbb{I}_{\{E_{t}(x; \theta_{t})<E_0\}}x}, \\
\end{aligned}
\end{equation}
where $S$ is the collection of selected reliable samples, $t$ represents the $t$-th iteration, and $\mathbb{I}$ is an indicator function. The model is updated by minimizing the entropy of reliable sample at the current time step, which is formulated as:

\begin{equation}\label{eq3}
\min_{\theta}{ 1 \over{M}}  \sum_{m=1}^M E(x_{m}) ,
\end{equation}
where $M$ denotes the number of selected reliable samples in $S_{(t;\theta_{t})}$.


Subsequently, the updated model predicts the same batch of test samples again. This cycle of model updating and sample selection repeats for a predefined number of iterations, a hyper-parameter as mentioned earlier. The iterative process will also terminate if the selected samples across two consecutive time steps are identical, as described by:
\begin{equation}\label{eq4}
\begin{aligned}
    &  S_{(t;\theta_{t})} == S_{(t-1;\theta_{t-1})}.  \\
\end{aligned}
\end{equation}

Before the next batch of samples arrives, the current batch will be discarded without being saved.

\subsection{Adaptively Integrating the Adapt Branch Network}
Based on the above description, potentially reliable samples will be explored and all of them will be employed to adjust the source model. However, achieving a balance between learning target-specific knowledge and domain-agnostic information in an unsupervised manner remains a challenge. Existing approaches either directly update all the normalization parameters~\cite{wang2021tent,niu2022efficient} or only update part of them~\cite{niu2023towards,lee2024entropy}. Such updates are insufficient to fully capture the target knowledge.


In a deep model, the shallow layers extract general features while the high layers generate domain-specific information~\cite{long2015learning}. To sufficiently utilize the discovered potentially reliable samples, we introduce an extra Adapt Branch network. The source model is segmented into two parts, i.e., the Shallow Layers that produce general features, and the Source Branch network that shares the same deep architecture and initialized parameters as the Adapt Branch network. All the parameters of the Source Branch network are frozen to preserve the source knowledge, while the normalization layers of the Adapt Branch network are updated to extract target-specific information. Notably, the introduction of the Adapt Branch network differs from the Teacher-Student scheme~\cite{hu2022teacher} and the Exponential Moving Average (EMA) model update~\cite{liang2022dine}, as the two networks in POEM collaborate to achieve better results—i.e., their outputs are combined to make final predictions. We will also discuss the impact of the structures of these networks in the next section.


In our proposed POEM approach, we denote the parameters of the Shallow Layers, the Source Branch network, and the Adapt Branch network as $G$, ${f}_{S}$, and ${f}_{A}$, respectively. The normalization process, akin to SAR\cite{niu2023towards}, is formulated as $\min{E(x;\theta')}$, where $\theta'$ denotes the affine parameters of all normalization layers in $G$ and ${f}_{A}$.


During the adaptation process, all the normalization layers of ${f}_{A}$ will be updated to promote the learning of target knowledge. For each test sample $x$, features $\varphi$ are first extracted through the Shallow Layers $G$. $\varphi$ serves as input to both the Source Branch and Adapt Branch networks, which separately generate two distinct and independent probability distributions, denoted as $\hat{y}_{g}$ and $\hat{y}_{a}$, respectively. These two outputs are combined as follows:
\begin{equation}\label{eq5}
   \hat{y} = \frac{\hat{y}_{g}  +  \hat{y}_{a}}{2},
\end{equation}
where $\hat{y}_{g}$ and $\hat{y}_{a}$ represent the source knowledge and the extracted target-specific information, respectively. The potentially reliable samples of this batch will be selected for the next round of update. Specifically, the pseudo-labels of these samples are utilized to update the Adapt Branch network. The Cross-Entropy (CE) loss will be used to guide the adaptation process, which is illustrated as:
\begin{equation}
\label{eq6} 
CE(p(x', \hat{y}) = - \sum_{c=1}^{C}\hat{y}_c \log(p_c(x')), 
\end{equation} 
where \(\hat{y}\) denotes the pseudo-labels assigned to each reliable sample $x'$ with one-hot probability distributions and $C$ denotes the number of categories. In the next section, we will experimentally validate the superiority of this loss function compared to the entropy minimization policy. 

\subsection{Distinctiveness of Potentially Reliable Samples}
\begin{figure}[t]
    \includegraphics[width=0.9\linewidth]{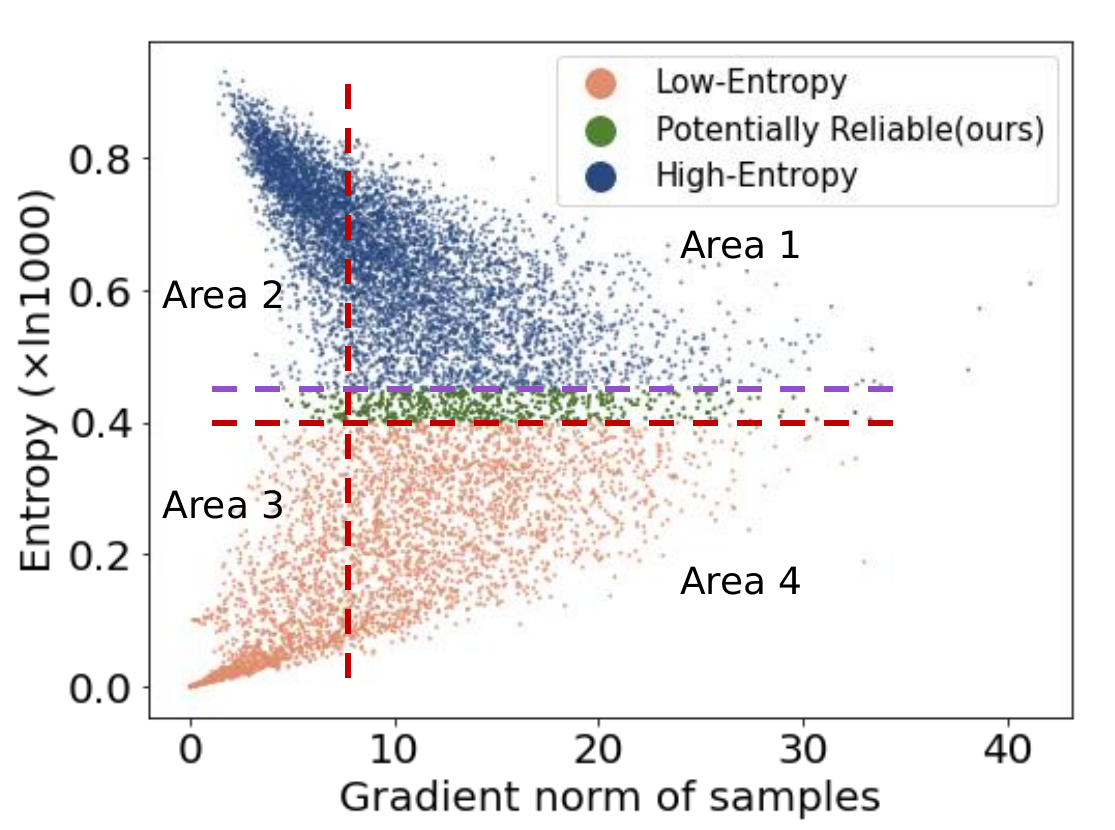}
    \caption{Samples in the green zone provide a normal range of gradients and high-quality supervised information for TTA on the dataset ImageNet-C. Therefore, it is necessary to utilize potentially reliable samples.}
    
    \label{gradnorm}
\end{figure}

An entropy-based TTA model is trained using an entropy-related loss function, with its parameters updated via gradients~\cite{wang2021tent}. In this unsupervised paradigm, the loss is derived from an entropy-related function, which depends on the self-supervised information and predicted labels of target samples. For example, both EATA~\cite{niu2022efficient} and DeYO ~\cite{lee2024entropy} select samples to promote the quality of supervised information for adaptation. Thus, the key question is whether potentially reliable samples can provide distinctive knowledge—namely, meaningful gradients and high-confidence pseudo-labels—to effectively adapt the model. As illustrated in Fig. \ref{gradnorm}, potentially reliable samples not only offer an appropriate range of gradients for adaptation but also provide high-quality pseudo-labels, i.e., valuable supervised information. In contrast, samples in Area 3, while having high-confidence pseudo-labels, fail to supply sufficient gradient information for model updates. Meanwhile, samples in Areas 1 and 2 lack high-quality pseudo-labels. Thus, identifying and leveraging previously unexplored potentially reliable samples is essential for effective test-time adaptation. \\

\begin{figure}[t]
    \includegraphics[width=0.85\linewidth]{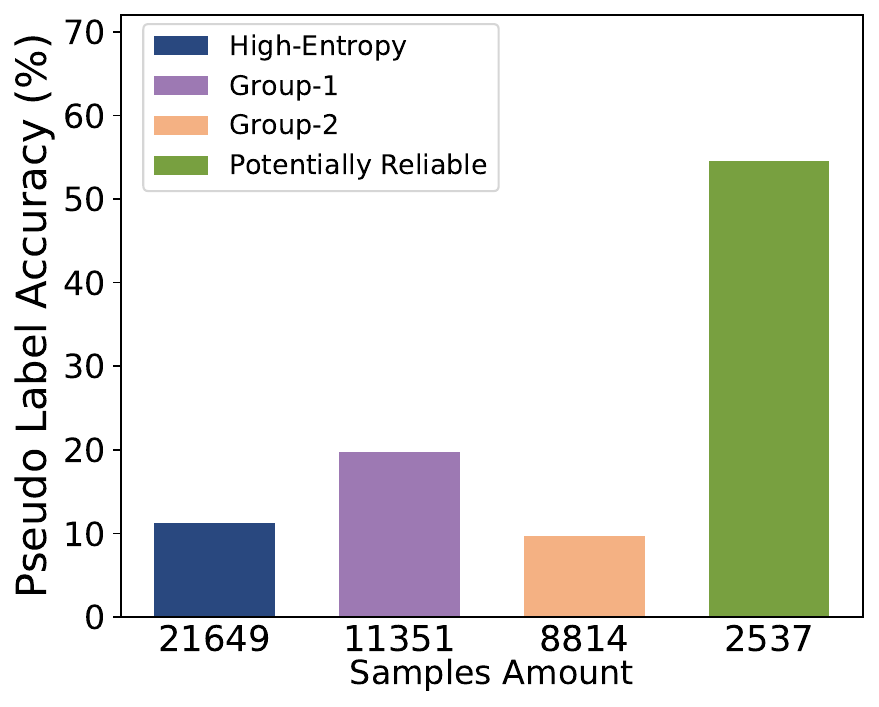}
    \caption{An illustration depicting the relationship between entropy range and the identification of potentially reliable samples. It is evident that such samples cannot be determined solely based on the entropy range. }
    \label{expentropyacc}
\end{figure}
\indent Building on the analysis above, another question arises: Is entropy the sole factor in identifying potentially reliable samples? Specifically, if two groups of samples share the same range of entropy, do they achieve the same pseudo-label accuracy? As shown in Fig. \ref{expentropyacc}, Group-1 represents all samples sharing the same entropy range as the potentially reliable samples, while Group-2 represents the samples obtained by excluding the potentially reliable samples from Group-1. We find that, despite sharing the same entropy range, the pseudo-label accuracy of the potentially reliable samples is significantly higher than that of both Group-1 and Group-2. This demonstrates that entropy alone is insufficient for identifying potentially reliable samples, underscoring their distinctive nature. Experimental results in the following section will further validate the importance of these samples. \\
\indent Based on the above analysis, we present the POEM algorithm in Algorithm~\ref{alg}. 

\begin{algorithm}[t]  
\caption{POEM Algorithm}  
\label{alg}  
\textbf{Input}: Test samples $D^{test}$. Source model ${f}_{\theta}$.   \\
\textbf{Output}: The predicted labels of $D^{test}$  
  
\begin{algorithmic}[1]  
\STATE Split ${f}_{\theta}$ into $G$ and ${f}_{S}$.
\STATE Deepcopy ${f}_{S}$ as ${f}_{A}$ . 
\STATE Collect the normalization layers $\theta$ and $\zeta$ from $G$ and ${f}_{A}$, respectively
\FOR{Each batch of samples $X$ in $D^{test}$}
    \STATE Calculate predictions $\hat{y}$ via  $G$, ${f}_{S}$ and ${f}_{A}$
    \STATE Compute entropy of $X$ based on Eqn.~\ref{eq1} 
    \STATE Select reliable samples $S$ based on Eqn.~\ref{eq2}
    \STATE Let $t=0$, $S'$=None
    \WHILE{$t \leq N$}  
        \IF{$S$ $\neq$ $S'$ }  
            \STATE Set $S'$=$S$
            \STATE Update $\theta$ of $G$ based on Eqn.~\ref{eq3}
            \STATE Update $\zeta$ of ${f}_{A}$ based on Eqn.~\ref{eq6}
            \STATE Re-predict $\hat{y}$ of the batch samples $X$
            \STATE Re-select reliable samples $S$.
            \STATE $t = t + 1$
        \ELSE
            \STATE Break
        \ENDIF    
        
    \ENDWHILE 
\ENDFOR  
\end{algorithmic}  
\end{algorithm}

\section{Experiments}
Our experiments focus on classification tasks designed to address several key questions: (1) Does POEM outperform the state-of-the-art (SOTA) entropy-based test-time adaptation (TTA) approaches? (2) Can POEM maintain its effectiveness in more challenging scenarios and datasets? (3) Can additional ablation studies confirm the effectiveness of POEM?

\subsection{Datasets and Baselines}
We conduct the comparative TTA experiments on two benchmark datasets: ImageNet-C and CIFAR100-C~\cite{hendrycks2019benchmarking}.  Each dataset contains corrupted images across 15 types of distortions in 4 main categories (noise, blur, weather, digital), with each type having 5 severity levels. We also explore real-world domain shift scenarios using the ImageNet-A~\cite{hendrycks2021natural}, ImageNet-R~\cite{hendrycks2021many}, and ImageNet-Sketch~\cite{wang2019learning} datasets to further validate the effectiveness of our proposed approach. ImageNet-R has renditions of 200 ImageNet classes resulting in 30,000 images, while ImageNet-Sketch consists of 50,889 images, approximately 50 images for each of the 1000 ImageNet classes. We compare POEM with several SOTA entropy-based TTA methods: Tent \cite{wang2021tent}, EATA \cite{niu2022efficient}, SAR \cite{niu2023towards}, DeYO \cite{lee2024entropy} and ROID~\cite{marsden2024universal}. Tent utilizes all test samples for adaptation. EATA filters out samples via entropy. SAR removes noisy gradients with large norms that may hurt the adaptation. DeYO introduces a new evaluation metric to measure reliable samples and assist the entropy-based approach. ROID is primarily based on the soft likelihood ratio loss and the symmetric cross-entropy loss. Additionally, we conduct ablation studies to evaluate the significance of each component in POEM.

\begin{table*}[t]
\centering
\setlength{\tabcolsep}{5pt} 
\caption{Comparative results based on ImageNet-C (\%). }
\label{ICmain}
\resizebox{\linewidth}{!}{
\begin{tabular}{l|ccc|cccc|cccc|cccc|c}
\multicolumn{1}{c}{} & \multicolumn{3}{c}{Noise} & \multicolumn{4}{c}{Blur} & \multicolumn{4}{c}{Weather} & \multicolumn{4}{c}{Digital} &  \\
\multicolumn{1}{c|}{Model / Method} & Gauss. & Shot & Impul. & Defoc. & Glass & Motion & Zoom & Snow & Frost & Fog & Brit. & Contr. & Elastic & Pixel & JPEG & Avg. \\ \midrule \midrule
\multicolumn{1}{c|}{ResNet-50-BN} & 2.9 & 3.6 & 2.6 & 17.8 & 9.7 & 14.7 & 22.4 & 16.6 & 23.0 & 23.9 & 59.1 & 5.3 & 16.5 & 20.8 & 32.6 & 18.1 \\ \midrule
\textbullet~Tent & 27.1 & 28.9 & 28.4 & 25.5 & 25.0 & 38.4 & 48.0 & 44.7 & 40.0 & 56.7 & 67.4 & 26.8 & 52.9 & 57.5 & 51.0 & 41.0 \\
\textbullet~EATA & 35.5 & \textbf{41.2} & \textbf{40.7} & 32.0 & 33.2 & 40.0 & 46.2 & 43.1 & 41.2 & 53.0 & 61.8 & 42.8 & 52.1 & 56.2 & 52.7 & 44.7 \\
\textbullet~SAR & 30.9 & 32.3 & 32.9 & 28.6 & 28.1 & 41.7 & 49.1 & 47.1 & 41.9 & 57.7 & 67.4 & 37.2 & 54.5 & 58.2 & 52.4 & 43.9 \\
\textbullet~DeYO & 35.5 & 37.4 & 36.9 & 33.5 & 32.9 & 46.8 & 52.5 & 51.6 & 45.8 & 60.0 & 68.6 & 42.4 & 58.0 & 60.9 & 55.5 & 47.8 \\
\textbullet~ROID & 29.1 & 30.8 & 30.1 & 26.7 & 26.4 & 40.9 & 48.7 & 47.6 & 40.1 & 57.0 & 66.9 & 36.6 & 54.5 & 57.7 & 51.4 & 43.0 \\
\textbf{\textbullet~POEM(ours)} & \textbf{36.5} & 38.9 & 38.2 & \textbf{34.6} & \textbf{34.2} & \textbf{48.3} & \textbf{53.4} & \textbf{52.6} & \textbf{46.7} & \textbf{60.6} & \textbf{68.8} & \textbf{46.2} & \textbf{58.6} & \textbf{61.7} & \textbf{56.0} & \textbf{49.0} \\ \midrule \midrule
\multicolumn{1}{c|}{ResNet-50-GN} & 22.0 & 23.0 & 22.0 & 19.7 & 11.4 & 21.4 & 25.0 & 40.2 & 46.9 & 34.0 & 68.8 & 36.2 & 18.5 & 29.2 & 52.6 & 31.3 \\ \midrule
\textbullet~Tent & 25.1 & 27.5 & 26.3 & 15.9 & 12.3 & 22.9 & 25.0 & 34.9 & 39.3 & 2.0 & 69.6 & 41.3 & 15.4 & 46.3 & 54.5 & 30.5 \\
\textbullet~EATA & 35.8 & \textbf{44.1} & \textbf{44.1} & 23.6 & \textbf{28.7} & 32.4 & 38.0 & 41.0 & 47.0 & 48.9 & 64.1 & 46.1 & 38.4 & 48.0 & 54.5 & 42.3 \\
\textbullet~SAR & 32.3 & 34.5 & 33.3 & 18.8 & 19.3 & 29.8 & 30.8 & 42.2 & 43.3 & 40.4 & 70.3 & 43.6 & 15.4 & 48.9 & 55.2 & 37.2 \\
\textbullet~DeYO & 36.9 & 38.9 & 37.7 & 20.0 & 22.0 & 34.0 & 33.4 & 47.7 & 46.3 & 4.1 & 71.9 & 47.2 & 33.8 & \textbf{52.9} & 56.6 & 38.7 \\
\textbf{\textbullet~POEM(ours)} & \textbf{40.3} & 42.3 & 41.9 & \textbf{30.5} & 27.4 & \textbf{38.7} & \textbf{39.8} & \textbf{53.5} & \textbf{51.2} & \textbf{58.5} & \textbf{73.6} & \textbf{55.1} & \textbf{43.7} & 52.4 & \textbf{57.9} & \textbf{47.2} \\ \midrule \midrule
\multicolumn{1}{c|}{ViTbase-LN} & 35.0 & 32.1 & 35.8 & 31.4 & 25.3 & 39.4 & 31.5 & 24.4 & 30.1 & 54.7 & 64.4 & 48.9 & 34.2 & 53.1 & 56.4 & 39.7 \\ \midrule
\textbullet~Tent & 51.7 & 51.4 & 52.9 & 51.4 & 47.6 & 55.9 & 49.1 & 10.7 & 18.0 & 66.8 & 73.2 & 66.4 & 52.0 & 65.6 & 64.4 & 51.0 \\
\textbullet~EATA & 53.6 & 56.8 & 57.2 & 52.9 & 57.0 & 60.5 & 60.2 & 63.6 & 64.5 & 70.0 & 77.5 & 64.5 & 67.0 & 71.0 & 69.0 & 63.0 \\
\textbullet~SAR & 49.8 & 49.5 & 50.9 & 48.2 & 42.4 & 49.9 & 45.5 & 17.4 & 46.2 & 63.2 & 72.9 & 63.5 & 45.4 & 63.3 & 61.2 & 51.2 \\
\textbullet~DeYO & 54.1 & 54.8 & 55.0 & 54.0 & 54.6 & 61.6 & 57.8 & 63.5 & 62.8 & 71.3 & 77.0 & 66.8 & 64.6 & 71.4 & 68.1 & 62.4 \\
\textbullet~ROID & 52.0 & 51.7 & 52.8 & 47.4 & 48.4 & 56.9 & 52.3 & 56.2 & 53.4 & 68.2 & 73.8 & 65.1 & 57.0 & 67.1 & 64.3 & 57.8 \\
\textbf{\textbullet~POEM(ours)} & \textbf{57.9} & \textbf{58.9} & \textbf{59.0} & \textbf{59.5} & \textbf{59.7} & \textbf{64.8} & \textbf{62.1} & \textbf{67.4} & \textbf{66.6} & \textbf{74.2} & \textbf{78.9} & \textbf{69.3} & \textbf{68.9} & \textbf{74.1} & \textbf{71.3} & \textbf{66.2} \\ \midrule
\end{tabular}
}
\end{table*}

\begin{table*}[h]
\centering
\caption{Comparative results based on  CIFAR100-C (\%).}
\label{ccmain}
\setlength{\tabcolsep}{5pt} 
\resizebox{\linewidth}{!}{
\begin{threeparttable}
\begin{tabular}{l|ccc|cccc|cccc|cccc|c}
 \multicolumn{1}{c}{}& \multicolumn{3}{c}{Noise} & \multicolumn{4}{c}{Blur} & \multicolumn{4}{c}{Weather} & \multicolumn{4}{c}{Digital} &  \\ 
\multicolumn{1}{c|}{Model / Method} & Gauss. & Shot & Impul. & Defoc. & Glass & Motion & Zoom & Snow & Frost & Fog & Brit. & Contr. & Elastic & Pixel & JPEG & Avg. \\ \midrule \midrule
\multicolumn{1}{c|}{ResNet-50-BN} & 40.2 & 40.8 & 37.8 & 59.5 & 48.8 & 58.3 & 61.5 & 50.8 & 51.2 & 51.1 & 61.8 & 54.0 & 54.7 & 56.2 & 46.4 & 51.5 \\ \midrule
\textbullet~Tent & 47.6 & 49.1 & 48.0 & 63.0 & 52.0 & 61.0 & 64.0 & 54.6 & 54.8 & 55.8 & 64.3 & 58.0 & 57.6 & 60.9 & 51.0 & 56.1 \\
\textbullet~EATA & 47.3 & 50.3 & 46.8 & 62.4 & 50.8 & 60.6 & 63.3 & 55.0 & 54.4 & 56.6 & 65.1 & 56.5 & 56.9 & 59.8 & 50.0 & 55.7 \\
\textbullet~SAR & 43.7 & 45.5 & 43.5 & 59.4 & 48.7 & 57.1 & 59.8 & 51.7 & 51.0 & 51.8 & 60.5 & 54.4 & 54.0 & 56.4 & 46.8 & 52.2 \\
\textbullet~DeYO & 47.7 & 50.3 & 48.5 & 65.4 & 54.6 & 64.0 & 67.1 & 57.1 & 57.1 & 57.6 & 67.3 & 63.1 & 60.7 & 63.2 & 52.6 & 58.4 \\
\textbullet~ROID & 36.2 &39.5 &33.4 &66.2 &39.3 &63.6 &68.0&\textbf{62.5}&62.2 &\textbf{64.8} &\textbf{75.8} &70.4 &51.7 &58.4 &42.4&55.6  \\
\textbf{\textbullet~POEM(ours)} & \textbf{55.4} & \textbf{58.6} & \textbf{55.8} & \textbf{70.2} & \textbf{60.4} & \textbf{67.9} & \textbf{70.5} & 61.8 & \textbf{62.5} & 63.8 & 71.4 & \textbf{67.9} & \textbf{65.1} & \textbf{67.1} & \textbf{58.4} & \textbf{63.8}\\ \midrule
\end{tabular}

\end{threeparttable}
}%

\end{table*}

\subsection{Implementation Details}
All experiments are implemented in Pytorch. The models are trained with 80 GB memory and an Nvidia 3090 GPU. We select ResNet-50-BN (Batch Normalization)~\cite{he2016deep}, ResNet-50-GN (Group Normalization), and ViTBase-LN (Layer Normalization)~\cite{dosovitskiy2021image} to thoroughly evaluate the effectiveness of POEM, as done similarly in SAR \cite{niu2023towards}. We use Stochastic Gradient Descent (SGD) with a momentum of 0.9. The batch size of target samples is 64. The learning rates of ResNet-based and ViT-based adaptation are set to 0.00025 and 0.001, respectively. The entropy threshold $E_0$, described in section 3, is set to 0.4× ln1000, following the methodology of EATA \cite{niu2022efficient}. As depicted in Fig.~\ref{overview}, only the normalization layers of the Shallow Layers and the Adapt Branch network are updated. All experiments using ImageNet-C and CIFAR100-C are conducted with samples at severity level 5, representing the most challenging conditions. The number of iterations, a critical hyper-parameter discussed in section 3.3, is set to 2. We will further explore the impact of this hyper-parameter later in this section. When integrating a baseline method with POEM, the entropy threshold remains as originally set in the baseline.

\subsection{Results}
Since POEM primarily consists of two strategies—potentially reliable samples and the Adapt Branch network—we will examine the importance of each strategy. POEM-I and POEM-II will be used to represent POEM without the Adapt Branch network and POEM without potentially reliable samples, respectively. Moreover, as POEM’s core concept can serve as an augmentation method to boost the performance of existing TTA approaches, we will evaluate the degree to which integrating these two key components enhances those approaches. All experiments focus on classification tasks, with performance measured in terms of classification accuracy (\%).\\

\subsubsection{Performance on ImageNet-C and CIFAR100-C} Comparative results based on these two datasets are displayed in Table \ref{ICmain} and Table \ref{ccmain}, respectively. POEM significantly outperforms all the original baseline approaches, demonstrating that the selection policy of potentially reliable samples and the Adapt Branch network can dramatically enhance TTA. Moreover, POEM consistently achieves the highest average performance, indicating its robustness to domain shifts.

\begin{table*}[t]
\setlength{\tabcolsep}{6pt} 
\centering
\caption{Performance under a challenging condition -- Imbalanced Label Shifts, i.e., online imbalanced label distribution shift, following the setting in SAR~\cite{niu2023towards} (ImageNet-C, \%). }
\label{ICLS}
\resizebox{\textwidth}{!}{

\begin{tabular}{l|ccc|cccc|cccc|cccc|c}
\multicolumn{1}{c}{}& \multicolumn{3}{c}{Noise} & \multicolumn{4}{c}{Blur} & \multicolumn{4}{c}{Weather} & \multicolumn{4}{c}{Digital} &  \\ 
\multicolumn{1}{c|}{Model / Method}& Gauss. & Shot & Impul. & Defoc. & Glass & Motion & Zoom & Snow & Frost & Fog & Brit. & Contr. & Elastic & Pixel & JPEG & Avg. \\ \midrule \midrule
\multicolumn{1}{c|}{ResNet-50-GN} & 22.0 & 23.0 & 22.0 & 19.7 & 11.4 & 21.4 & 25.0 & 40.2 & 46.9 & 34.0 & 68.8 & 36.2 & 18.5 & 29.2 & 52.6 & 31.3 \\ \midrule
\textbullet~  Tent & 20.8 & 23.1 & 23.9 & 13.8 & 8.6 & 19.6 & 20.6 & 26.7 & 33.2 & 9.3 & 69.8 & 42.2 & 13.1 & 49.5 & 53.6 & 28.5 \\
\textbullet~  EATA & 27.7 & 28.7 & 28.1 & 15.2 & 19.3 & 24.9 & \textbf{27.9} & 32.3 & 25.4 & 37.4 & 64.1 & 34.1 & 26.4 & 43.4 & 45.2 & 32.0 \\
\textbullet~  SAR & 35.5 & 37.0 & 36.2 & 18.8 & 9.8 & 32.7 & 17.2 & 22.5 & 44.8 & \textbf{47.6} & 71.0 & 45.9 & 7.0 & 51.2 & 55.8 & 35.5 \\
\textbullet~  DeYO & 43.0 & 44.9 & \textbf{43.8} & 23.0 & 19.4 & \textbf{40.5} & 3.6 & 51.6 & \textbf{51.5} & 12.3 & \textbf{73.2} & \textbf{52.5} & 10.2 & 58.6 & 58.9 & 38.4 \\
 
\textbf{\textbullet~  POEM(ours)} & \textbf{43.4} & \textbf{45.0} & \textbf{43.8} & \textbf{24.1} & \textbf{20.0} & 40.1 & 3.8 & \textbf{52.0} & \textbf{51.5} & 17.0 & \textbf{73.2} & \textbf{52.5} & \textbf{46.6} & \textbf{59.0} & \textbf{59.0} & \textbf{41.0} \\ \midrule \midrule
\multicolumn{1}{c|}{ViTbase-LN} & 35.0 & 32.1 & 35.8 & 31.4 & 25.3 & 39.4 & 31.5 & 24.4 & 30.1 & 54.7 & 64.4 & 48.9 & 34.2 & 53.1 & 56.4 & 39.7 \\ \midrule
\textbullet~  Tent & 55.1 & 54.3 & 55.9 & 56.2 & 54.2 & 59.9 & 48.5 & 5.8 & 16.3 & 70.8 & 75.5 & 68.8 & 60.9 & 69.8 & 67.2 & 54.6 \\
\textbullet~  EATA & 43.8 & 39.9 & 35.1 & 19.8 & 26.6 & 21.1 & 22.1 & 21.4 & 24.9 & 18.4 & 42.3 & 4.3 & 26.8 & 24.3 & 27.4 & 26.5 \\
\textbullet~  SAR & 52.9 & 53.1 & 54.1 & 52.7 & 52.9 & 57.4 & \textbf{52.8} & 8.4 & 31.3 & 68.2 & 74.5 & \textbf{66.6} & 58.8 & 67.1 & 65.0  & 54.5 \\
\textbullet~  DeYO & 53.9 & 54.7 & 54.8 & 53.8 & 55.5 & 61.0 & 3.5 & 64.8 & 63.1 & 71.0  & 77.0  & 64.6 & 67.3 & 71.8 & 68.4 & 59.0 \\
 
\textbf{\textbullet~  POEM(ours)} & \textbf{56.8} & \textbf{57.6} & \textbf{57.6} & \textbf{57.6} & \textbf{59.5} & \textbf{65.0} & 9.8 & \textbf{68.8} & \textbf{66.3} & \textbf{73.9} & \textbf{78.7} & 65.9 & \textbf{70.1} & \textbf{73.7} & \textbf{71.6} & \textbf{62.2} \\ \midrule
\end{tabular}%
}
\end{table*}

\begin{table}[t]
\centering
\caption{Performance under the setting of mixed domain shifts~\cite{niu2023towards} (ImageNet-C, \%). }
\label{mixshifts}
\setlength{\tabcolsep}{10pt} 
\resizebox{0.8\linewidth}{!}{%
\begin{tabular}{ccc}
\multicolumn{1}{c}{}& \multicolumn{1}{c}{ResNet-50-GN} &  ViTbase-LN \\ \midrule \midrule 
 Tent &  33.4 &  24.1 \\
 EATA &  38.1 &  56.4 \\
 SAR &  38.6 &  56.6 \\
 DeYO &  39.1 &  57.8 \\ 
\multicolumn{1}{c}{\textbf{ POEM(ours)}} & \textbf{40.1} & \textbf{58.3} \\ \midrule
\end{tabular}%
}

\end{table}

\begin{table}[t]
\centering
\caption{Comparative results based on real-world datasets, i.e., ImageNet-sketch, ImageNet-R, and ImageNet-A (\%).}
\label{otherds}
\setlength{\tabcolsep}{4pt} 
\resizebox{\linewidth}{!}{%
\begin{tabular}{ccccc}
 \multicolumn{1}{c}{}& \multicolumn{1}{c}{ImageNet-A} & \multicolumn{1}{c}{ImageNet-R} & \multicolumn{1}{c}{ImageNet-sketch} & Avg. \\ \midrule \midrule
Tent & 52.0 & 41.9 & 30.4 & 41.4 \\
EATA & 52.6 & 44.3 & 35.1 & 44.0 \\
SAR & 51.7 & 42.7 & 31.6 & 42.0 \\
DeYO & 53.2 & 45.2 & 35.5 & 44.6 \\ 
\textbf{POEM(ours)} & \textbf{54.2} & \textbf{46.4} & \textbf{35.9} & \textbf{45.5} \\ \midrule
\end{tabular}%
}
\end{table}

\subsubsection{Performance on more challenging scenarios} The settings described in to Table \ref{ICmain} and Table \ref{ccmain} represent standard conditions. To assess POEM's performance under more challenging conditions, we explore three additional scenarios, as outlined in SAR \cite{niu2023towards}. (1) Online Imbalance Label Distribution. This scenario involves dynamic shifts in the ground-truth test label distribution, leading to imbalanced distributions for each corruption type, where the class imbalance ratio approaches $\infty$. Table~\ref{ICLS} demonstrates that POEM outperforms all SOTA methods, achieving at least 3.2\% average improvement with ViT-Base model. (2) Single-Sample Arrival. The target samples arrive one by one, i.e., the batch size is set to 1. Table~\ref{ICbs1} illustrates that POEM also leads over all other methods, providing a minimum of a 1\% improvement. Moreover, POEM shows more than 1.0\% improvement in average over DeYO. (3) Mixed Domain Shifts. This scenario involves the combination of multiple domains, following the setting in SAR~\cite{niu2023towards} (\%). As shown in Table~\ref{mixshifts}, POEM surpasses all SOTA methods, delivering approximately a 1.5\% improvement over the best-performing SOTA methods. Moreover, POEM shows an improvement of more than 1.5\% over DeYO. Overall, POEM consistently outperforms all baseline approaches in these more challenging scenarios, demonstrating its robustness and effectiveness.

\subsubsection{Performance on more challenging datasets} 

We have also carried out additional experiments on more challenging, real-world datasets, including ImageNet-A (ViT-Base), ImageNet-R (ResNet-50-BN), and ImageNet-Sketch (ResNet-50-BN). The results, presented in Table \ref{otherds}, demonstrate that POEM continues to outperform the comparative methods, further confirming its robustness in real-world domain shifts.

\begin{table}[t]
\centering 
\caption{Significance of exploring potentially reliable samples (ImageNet-C, \%).} 
\label{KeyAdvantage}
\setlength{\tabcolsep}{6pt} 
\resizebox{\linewidth}{!}{
\begin{tabular}{lccccc}
\multicolumn{1}{c}{Method} & Gauss. & Defoc. & Snow & \multicolumn{1}{c}{Contr.} & \multicolumn{1}{c}{Avg.} \\ \midrule \midrule
Tent* & 29.0 & 27.1 & 45.7 & 31.4 & 33.3 \\
Tent + POEM-I & 30.3 & 28.4 & 45.8 & 34.6 & \multicolumn{1}{l}{34.8} \\
EATA* & 35.2 & 32.6 & 51.6 & 43.9 & 40.7 \\
EATA + POEM-I & 35.6 & 33.3 & 51.8 & 44.3 & 41.3 \\
SAR* & 32.4 & 30.1 & 48.5 & 40.1 & 37.7 \\
SAR + POEM-I & 34.2 & 31.3 & 49.8 & 40.2 & 38.9 \\
DeYO* & 36.0 & 33.8 & 52.0 & 44.0 & 41.4 \\
DeYO + POEM-I & \textbf{36.2} & \textbf{34.0} & \textbf{52.4} & \textbf{45.6} & \textbf{42.1}\\ \midrule
\end{tabular}
}

\end{table}

\begin{table*}[t]
\caption{Performance under a challenging condition -- the batch size is set to 1, following the setting in SAR~\cite{niu2023towards} (ImageNet-C, \%). }
\label{ICbs1}
\centering
\setlength{\tabcolsep}{5pt} 
\resizebox{\textwidth}{!}{
\begin{tabular}{l|ccc|cccc|cccc|cccc|c}
\multicolumn{1}{c}{}& \multicolumn{3}{c}{Noise} & \multicolumn{4}{c}{Blur} & \multicolumn{4}{c}{Weather} & \multicolumn{4}{c}{Digital} &  \\ 
\multicolumn{1}{c|}{Model / Method}& Gauss. & Shot & Impul. & Defoc. & Glass & Motion & Zoom & Snow & Frost & Fog & Brit. & Contr. & Elastic & Pixel & JPEG & Avg. \\ \midrule \midrule
\multicolumn{1}{c|}{ResNet-50-GN} & 18.0 & 19.8 & 17.9 & 19.8 & 11.4 & 21.4 & 24.9 & 40.4 & 47.3 & 33.6 & 69.3 & 36.3 & 18.6 & 28.4 & 52.3 & 30.6 \\ \midrule
\textbullet~Tent & 30.8 & 35.7 & 33.6 & 16.7 & 14.2 & 27.2 & \textbf{31.3} & 18.8 & 26.5 & 2.1 & 71.8 & 46.2 & 7.0 & 52.5 & 56.3 & 31.3 \\
\textbullet~EATA & 29.3 & 39.4 & 38.9 & 19.7 & 18.2 & 22.3 & 26.8 & 38.6 & 42.2 & \textbf{40.3} & 62.8 & 36.6 & 24.4 & 38.7 & 49.0 & 35.1 \\
\textbullet~SAR & 28.4 & 30.8 & 29.4 & 18.5 & 15.1 & 28.7 & 30.6 & 45.1 & 44.1 & 32.6 & 71.9 & 44.6 & 12.3 & 47.7 & 56.1 & 35.7 \\
\textbullet~DeYO & 41.2 & 44.5 & 43.2 & \textbf{22.6} & \textbf{24.4} & 40.6 & 5.4 & \textbf{53.4} & \textbf{51.5} & 3.1 & 73.1 & 52.8 & 47.6 & 59.2 & \textbf{59.4} & 41.4 \\
\textbf{\textbullet~POEM(ours)} & \textbf{42.8} & \textbf{45.3} & \textbf{44.4} & 21.9 & 19.1 & \textbf{41.6} & \textbf{32.2} & 53.0 & 46.5 & 1.5 & \textbf{73.4} & \textbf{53.1} & \textbf{47.6} & \textbf{60.3} & 59 & \textbf{42.8} \\ \midrule \midrule
\multicolumn{1}{c|}{ViTbase- LN} & 9.5 & 6.8 & 8.2 & 29.0 & 23.5 & 33.9 & 27.1 & 15.9 & 26.5 & 47.2 & 54.7 & 44.1 & 30.5 & 44.5 & 47.8 & 29.9 \\ \midrule
\textbullet~Tent & 51.9 & 51.7 & 53.0 & 52.2 & 48.3 & 56.5 & 48.9 & 8.1 & 13.1 & 67.4 & 73.4 & 66.5 & 52.7 & 64.7 & 63.9 & 51.4 \\
\textbullet~EATA & 47.7 & 52.2 & 56.3 & 50.1 & 50.6 & 56.8 & 52.4 & 49.8 & 58.7 & 67.7 & 76.5 & 65.6 & 59.8 & 68.9 & 69.3 & 58.8 \\
\textbullet~SAR & 51.8 & 51.3 & 53.0 & 50.6 & 48.9 & 57.0 & 50.9 & 19.9 & 55.2 & 68.2 & 74.9 & 65.8 & 54.3 & 66.5 & 64.7 & 55.5 \\
\textbullet~DeYO & 53.5 & 55.2 & 55.6 & 54.8 & 54.9 & 62.5 & \textbf{55.4} & \textbf{64.7} & \textbf{63.6} & 71.6 & \textbf{77.3} & 67.2 & \textbf{65.8} & 71.6 & 68.7 & 62.8 \\
\textbf{\textbullet~POEM(ours)} & \textbf{56.3} & \textbf{56.8} & \textbf{57.9} & \textbf{58.0} & \textbf{57.8} & \textbf{63.4} & 53.6 & 63.2 & 62.5 & \textbf{73.0} & \textbf{77.3} & \textbf{69.7} & 65.1 & \textbf{72.4} & \textbf{69.7} & \textbf{63.8} \\ \midrule
\end{tabular}%
}
\end{table*}

\subsection{Analysis}
\subsubsection{Importance of potentially reliable samples} (1) Potentially Reliable Samples \emph{\underline{VS}} Adapt Branch network. We assess whether POEM-I and POEM-II can enhance existing approaches independently. The comparative results, shown in Table \ref{POA_POB}, reveal that both POEM-I and POEM-II improve existing techniques, with POEM-I consistently outperforming POEM-II. This suggests that leveraging potentially reliable samples is more beneficial than relying solely on the Adapt Branch network, which helps balance learning new knowledge with retaining existing information. (2) Potentially Reliable Samples \emph{\underline{VS}} Iterative Update. We investigate whether increasing the number of model updates improves the performance. As shown in Table \ref{KeyAdvantage}, an asterisk (*) indicates the original method with iterative updating. In this comparison, all approaches undergo the same number of iterations. Despite this, POEM consistently outperforms these SOTA methods, highlighting that the primary advantage of POEM lies in its exploration of potentially reliable samples rather than in the iterative update process. Therefore, identifying these samples is more crucial than iterative updating.

\subsubsection{Sensitivity to the predefined entropy threshold} 
\begin{figure}[t]
    \centering  %
    \includegraphics[width=0.75\linewidth]{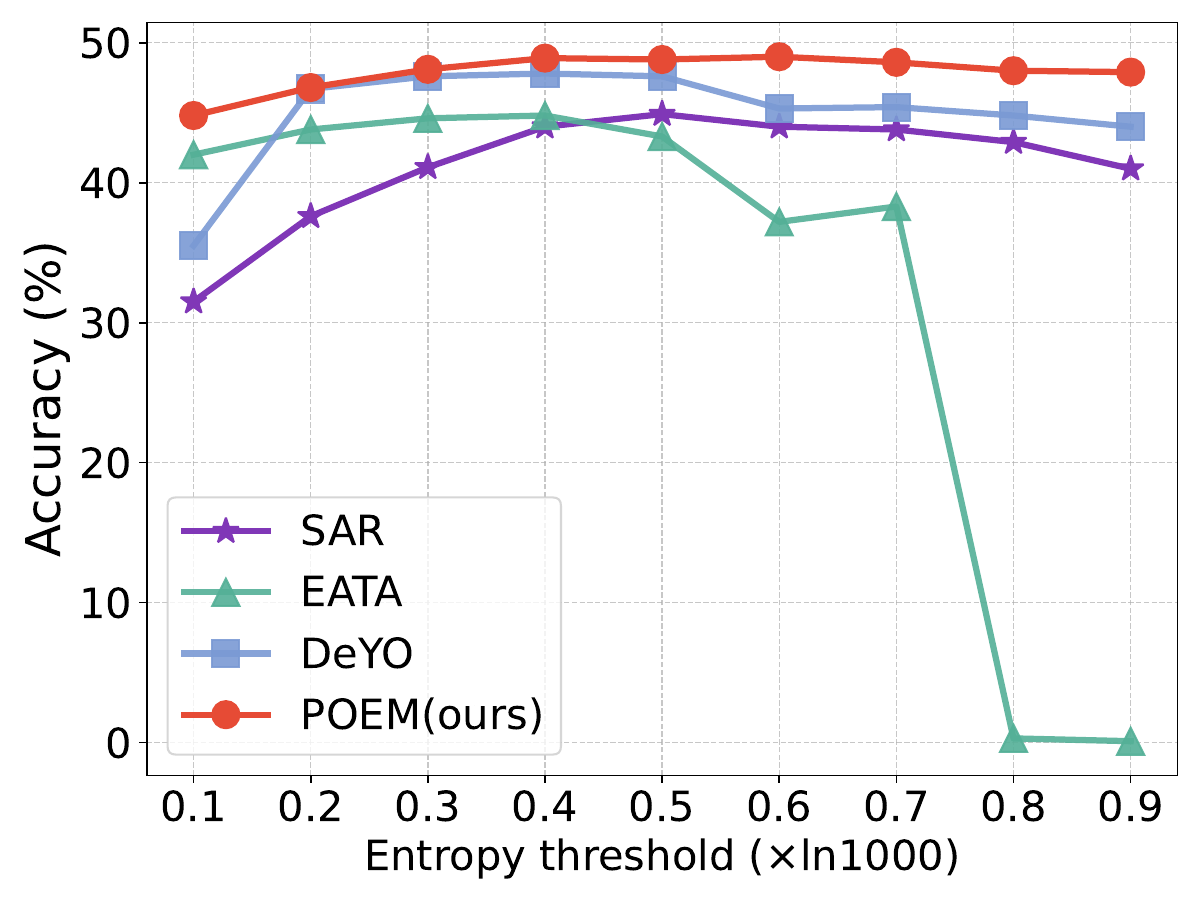}
    \caption{Sensitivity to the predefined entropy threshold (ImageNet-C). POEM demonstrates robustness to variations in the threshold and consistently outperforms SOTA approaches.}
    
    \label{threshold}
\end{figure}

To further understand the robustness of POEM, we explore its sensitivity to the predefined entropy threshold, which dictates the selection of samples for model updates. The comparative results are displayed in Fig.~\ref{threshold}. POEM exhibits greater stability compared to the baselines, demonstrating much less sensitivity to the entropy threshold. We observe that when the threshold is set above 0.8, the performance of EATA~\cite{niu2022efficient} drops dramatically. This decline is attributed to EATA’s re-weighting mechanism, which assigns higher weights to high-entropy samples, some of which may have incorrect labels, thereby negatively impacting the training process.

\begin{table*}[t]
\centering
\caption{Comparison of the importance of POEM-I and POEM-II as general augmentation strategies (ImageNet-C, \%). }
\setlength{\tabcolsep}{4pt} 
\label{POA_POB}
\resizebox{\textwidth}{!}{%
\begin{tabular}{c|ccc|ccc|ccc|ccc}
Model / Method & Tent & + POEM-I & + POEM-II & EATA & + POEM-I & + POEM-II & SAR & + POEM-I & + POEM-II & DeYO & + POEM-I & + POEM-II \\ \midrule \midrule
ResNet-50-BN & 41.2 & \textbf{43.1} & 42.6 & 44.7 & \textbf{46.3} & 46.1 & 43.9 & \textbf{46.3} & 44.1 & 47.8 & \textbf{48.6} & \textbf{48.6} \\
ResNet-50-GN & 30.5 & \textbf{31.2} & 31.0 & 42.3 & \textbf{43.4} & 43.0 & 36.9 & \textbf{37.5} & 37.4 & 38.7 & \textbf{41.1} & 40.3 \\
ViTbase-LN & 51.7 & \textbf{52.9} & 52.8 & 63.0 & \textbf{64.9} & 63.2 & 51.2 & \textbf{56.4} & 55.2 & 62.4 & \textbf{64.3} & 64.1\\ \midrule
\end{tabular}
}
\end{table*}

\begin{figure}[t]
    \centering  
    \includegraphics[width=\linewidth]{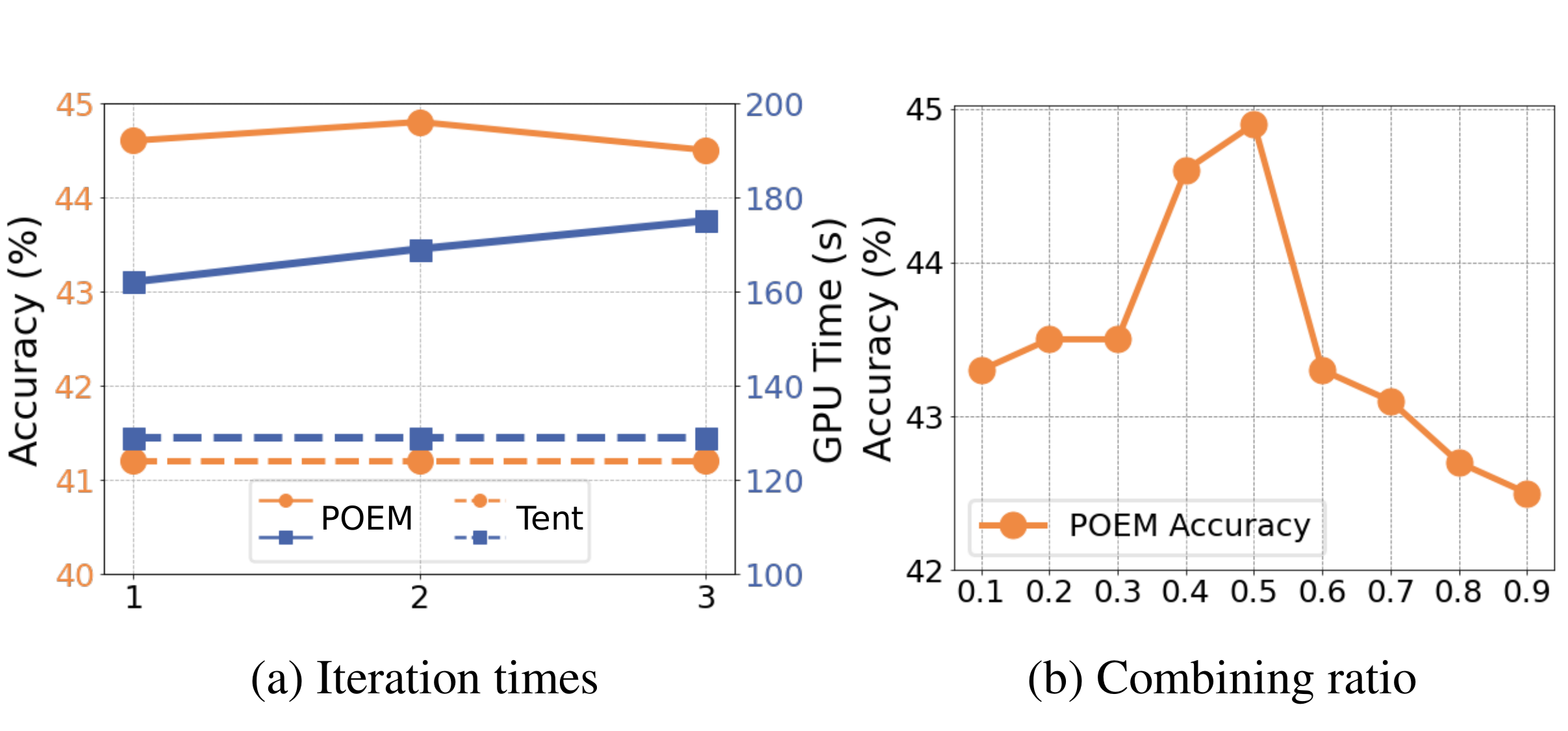}
    \caption{Sensitivity of hyper-parameters (ImageNet-C). (a) Adding one extra iteration is sufficient for optimal performance, and adding a second extra iteration provides only marginal improvement. (b) The optimal ratio for combining the outputs of the two networks is 1:1.}
    \vspace{-0.1in}
    \label{hyper}
\end{figure}

\subsubsection{Discussion of hyperparameters, including iteration count and combination ratio} (1) Iteration Count. In POEM, each additional iteration represents a model update using the discovered potentially reliable samples. The comparative results for the hyper-parameter of iteration count for model updates are shown in Fig.~\ref{hyper}(a). It is evident that adding one extra iteration is sufficient, while two additional iterations only provide a marginal improvement. In fact, more iterations can even reduce performance. This is because POEM effectively identifies potentially reliable samples within a limited number of rounds. Compared to Tent~\cite{wang2021tent}, POEM achieves significantly higher performance without a substantial increase in GPU time. (2) Combining Ratio. The impact of the ratio for combining the outputs of the two networks (as defined in Eqn. \ref{eq5}) is illustrated in Fig.~\ref{hyper}(b). This Equation is rewritten as $\hat{y}$=$\alpha$*$\hat{y}_{g}$  +  (1-$\alpha$)*$\hat{y}_{a}$). The outputs of the source model and the Adapt Branch network are combined to produce final predictions.  The results show that POEM achieves the best performance with a 1:1 ratio. This suggests that balancing the learning of new knowledge with retaining existing information is crucial for optimal performance.

\begin{table*}[!t]
\caption{Characteristics of POEM (Gaussian noise of ImageNet-C). }
\label{eff}
\resizebox{\textwidth}{!}{
\begin{tabular}{c|cc|ccccc}
& Source Data? & Online? & \#Forward & \#Backward & Other Computations & GPU Time (s)& FPS\\ \midrule \midrule
No Adapt. & $\times$ & $\times$ & 50,000 & - & N/A & 94&531 \\
MEMO & $\times$ & $\times$ & 50,000$\times$65 & 50,000$\times$64 & Augmentation\&Mix & 45,510 &1.1\\
Tent & $\times$ & $\checkmark$ & 50,000 & 50,000 & N/A & 127 &393\\
EATA & $\checkmark$ & $\checkmark$ & 50,000 & 19,085 & Regularizer & 153 &326\\
SAR & $\times$ & $\checkmark$ & 50,000 + 18,608 & 19,030 + 12,491 & Additional model update & 181 &276\\
DeYO & $\times$ & $\checkmark$ & 50,000 + 33,943 & 31,806 & Additional PLPD computing & 165 &303\\ 
\textbf{POEM(ours)} & $\times$ & $\checkmark$ & 50,000$\times$2 & 41,608 & Additional parameters update & 238&210\\ \midrule
\end{tabular}%
}
\end{table*}

\begin{table}[t]
\centering
\caption{Impact of different architectures for the Adapt Branch network (ImageNet-C, \%). }
\label{Adapt_Full}
    \centering
    \resizebox{\linewidth}{!}{
    \begin{tabular}{c|cccc|c}
 \multicolumn{1}{c}{}& \multicolumn{1}{c}{Gauss.} & \multicolumn{1}{c}{Defoc.} & \multicolumn{1}{c}{Snow} & \multicolumn{1}{c}{Contr.} & Avg. \\ \midrule \midrule
ViTbase-LN & 55.8 & 57.0 & 61.8 & 68.5 & 46.1 \\ \midrule
Block 11 & 56.6 & 57.7 & 62.5 & 69.2 & 61.5 \\
Block 10+11 & 56.6 & 57.8 & 62.5 & 69.2 & 61.5 \\
Block 9+10+11 & 56.6 & \textbf{57.9} & 62.6 & \textbf{69.3} & 61.6 \\ \midrule \midrule
ResNet-50 & 29.6 & 27.7 & 47.0 & 25.3 & 32.4 \\ \midrule
Layer 4 & \textbf{31.8} & \textbf{29.7} & \textbf{47.6} & \textbf{35.6} & \textbf{36.2} \\
Layer 3+4 & 31.7 & 28.0 & 47.3 & 34.4 & 35.4 \\
Layer 2+3+4 & 30.6 & 28.5 & 46.7 & 35.3 & 35.2\\ \midrule
\end{tabular}
    }
\end{table}

\begin{table}[t]  
\centering  
\caption{Comparison of updating different normalization layers based on Tent~\cite{wang2021tent} (ImageNet-C, \%).  }
\label{Tent_Shallow_All}  

\resizebox{\linewidth}{!}{
\setlength{\tabcolsep}{4pt} 
    \renewcommand{\arraystretch}{1.3} 
\begin{tabular}{c|c|cccc|c}
\multicolumn{1}{c}{}&\multicolumn{1}{c}{Layers to update} & \multicolumn{1}{c}{Gauss.} & Defoc. & Snow & \multicolumn{1}{c}{Contr.} & \multicolumn{1}{c}{Avg.} \\ \midrule \midrule
ResNet-50 & Only shallow layers & 27.1 & 25.5 & 45.2 & 28.0 & 31.4 \\
 & All layers & 27.1 & 25.5 & 44.7 & 26.8 & 31.0 \\ \midrule
ViTbase-LN & Only shallow layers & 51.8 & 51.3 & 10.8 & 66.3 & 45.0 \\
\multicolumn{1}{l|}{} & All layers & 51.7 & 51.4 & 10.7 & 66.4 & 45.0 \\ \midrule
\end{tabular}
}
\end{table}

\subsubsection{Design of the Adapt Branch network} Table~\ref{Tent_Shallow_All} shows that applying uniform normalization operations across the entire network may not be the optimal solution. Consequently, the next challenge is to design the Adapt Branch network. Sharing the same initialization as the Source Branch network, it plays a crucial role in the final feature extraction. Comparative results are presented in Table~\ref{Adapt_Full}, where experiments are conducted using Tent~\cite{wang2021tent} and ImageNet-C. For a ResNet-based~\cite{he2016deep} deep model, assigning the last convolutional layer as the Adapt Branch network achieves the best performance. This suggests that updating the normalization parameters of the highest layer is sufficient to extract target-specific knowledge. Similarly, for a ViT-based~\cite{dosovitskiy2021image} architecture, updating the normalization layer parameters of the last block yields approximately the best performance.

\subsubsection{The influence of the loss function for the Adapt Branch network}
We select different loss functions for the Source Branch network and the Adapt Branch network, as shown in Fig.~\ref{overview}. The parameters of the Source Branch network are fixed to preserve existing knowledge, while the parameters of the Adapt Branch network are updated to adapt to new data. Since the higher layers of deep networks are highly sensitive, the parameters of the Adapt Branch network are particularly delicate. Therefore, the supervised information provided to this network must be extremely precise. Consequently, we choose cross-entropy as the loss function, instead of entropy minimization. Our additional comparative experiments reveal that the performance of POEM decreases from 36.8\% to 36.3\% when the loss function is switched from cross-entropy to entropy minimization. This demonstrates the advantage of using cross-entropy as the loss function in this context. The likely reason is that cross-entropy focuses on optimizing a specific category, while entropy minimization requires considering all possible categories. As a result, cross-entropy is more effective in guiding model adaptation compared to the entropy-based policy.

\begin{table}[t]
\centering
\caption{Performance comparison when baseline methods integrate with POEM's two components, i.e., potentially reliable samples, and the Adapt Branch network (ImageNet-C, \%).}
\label{integrate}
\setlength{\tabcolsep}{5pt} 
\resizebox{0.95\linewidth}{!}{%
\begin{tabular}{lcccc}
\multicolumn{1}{c}{} & ResNet-50-BN & ResNet-50-GN & ViT-Base-LN & Avg. \\ \midrule \midrule
Tent & { 41.0} & { 30.5} & { 51.0} & { 40.8} \\
+ POEM & { 44.9} & { 34.3} & { 53.6} & { 44.3} \\ \midrule
EATA & { 44.7} & { 42.3} & { 63.0} & { 50.0} \\
+ POEM & { 46.9} & { 44.5} & { 65.9} & { 52.4} \\ \midrule
SAR & { 43.9} & { 37.2} & { 51.2} & { 44.1} \\
+ POEM & { 46.7} & { 42.0} & { 59.7} & { 49.5} \\ \midrule
DeYO & { 47.8} & { 38.7} & { 62.4} & { 49.6} \\
+ POEM & { 48.8} & { 43.0} & { 65.5} & { 52.4}\\ \midrule
\end{tabular}
}
\end{table}

\subsubsection{Can POEM work in real-time?} We compare the computational efficiency of various TTA approaches, with results shown in Table \ref{eff}.   We also provide computational details for different TTA approaches using ResNet-50-BN and the ImageNet-C dataset (50,000 images). POEM does not require significantly more computational time compared to other methods. In contrast, MEMO~\cite{zhang2022memo} incurs a much higher computational cost due to the need to update all parameters. Specifically, POEM can process 200 images per second (Frame Per Second, FPS). Therefore, POEM is suitable for real-time applications and is potentially applicable in dynamic scenarios such as autonomous driving.

\subsubsection{Can the key components of POEM be integrated into existing methods?} Since both components of POEM are general policies, we investigate whether they can be incorporated into existing approaches to enhance overall performance. In each combination (+POEM), the integrated version retains POEM’s deep architecture while preserving the original sample selection policy, such as EATA. The results, presented in Table~\ref{integrate}, clearly demonstrate that the key components of POEM consistently enhance performance, establishing it as an effective general augmentation strategy. Therefore, POEM can serve as a versatile framework for improving future entropy-based TTA techniques.

\section{Conclusions}
In this paper, we propose POEM, a novel approach that enhances test-time adaptation (TTA) by exploring potentially reliable target samples. POEM identifies them through iterative entropy minimization, enabling initially overlooked samples to surpass the threshold and become reliable for adaptation. These samples provide a normal range of gradients and high-confident supervised information to effectively update the model. Furthermore, we introduce an additional Adapt Branch network designed to capture target-specific knowledge. By integrating its outputs with those of the source model, POEM achieves a balance between target-specific learning and the extraction of domain-agnostic representations, making it well-suited for complex or unpredictable scenarios.

Extensive experiments demonstrate that POEM not only outperforms state-of-the-art entropy-based approaches but also enhances existing techniques as a general augmented strategy. Its effectiveness is validated in both corruption and real-world domain shift scenarios. Unlike existing entropy-based methods, POEM shows robustness to the choice of entropy threshold. Furthermore, the exploration of potentially reliable samples does not significantly increase computational cost, allowing POEM to operate in real-time. 



\bibliographystyle{ieeetr} 
\bibliography{references} 

\end{document}